\documentclass[11pt]{article}

\usepackage[preprint]{acl}

\usepackage{times}
\usepackage{latexsym}
\usepackage[T1]{fontenc}
\usepackage[utf8]{inputenc}
\usepackage{microtype}
\usepackage{inconsolata}
\usepackage{graphicx}
\usepackage{subcaption}
\usepackage{amsmath}
\usepackage{booktabs}
\usepackage{fancyvrb}

\title{Recognizing Atari and Assessing Sente: Controlled Evaluation Tasks for Language-Based Go Agents}

\author{
  Taiyi Chen\thanks{Equal contribution.},
  Iskar Deng\footnotemark[1],
  Amandalynne Paullada\\
  University of Washington\\
  \texttt{\{taiyic2,hd49,paullada\}@uw.edu}
}

\begin{document}
\maketitle

\begin{abstract}
Domain terminology provides concise descriptions of game states and strategic relationships, but whether game-playing language models can accurately recognize these concepts and relate them to move evaluation remains unclear. We construct a controlled evaluation set centered on \textit{atari} and its relationship to \textit{sente} in Go, examining terminology recognition under local board changes and move evaluation under different strategic conditions. We conduct our tests on LoGos, a Go language model, after fine-tuning it with terminology and commentary supervision. The model struggles to recognize changes in the \textit{atari} relationship and assess its strategic implications. Its terminology predictions are associated with move preferences even when those preferences contradict the reference evaluation. Our controlled tests provide a framework for examining whether language-based Go agents connect domain terminology with precise board relationships and strategic move values.
\end{abstract}

\section{Introduction}
\label{sec:introduction}

In board games, players use domain terminology to describe game states, explain moves, and discuss tactics and strategies \citep{jhamtani-etal-2018-learning,zang-etal-2019-automated}.
Classical studies of chess expertise suggest that experts organize and retrieve information about board positions through meaningful patterns of pieces, chunks, and templates \citep{chase-simon-1973-perception,gobet-simon-1996-templates}.
Domain terminology provides names for recurring patterns and relationships on the board, allowing players to describe and communicate game knowledge in concise language \citep{tomlin-etal-2022-understanding}.

Research on game-playing AI has shown that models can develop internal representations that align with human concepts. \citet{mcgrath-etal-2022-acquisition} find that representations of several human chess concepts gradually emerge during AlphaZero's training. \citet{tomlin-etal-2022-understanding} further show that intermediate representations of Go policy networks can be used to predict mentions of domain terminology in human commentary. However, the ability to decode concept information from a model's internal representations does not necessarily imply that the model can appropriately apply those concepts in its actions. \citet{lovering-etal-2022-evaluation} analyze a Hex-playing agent using concept probing and targeted behavioral tests, finding that it encodes information about strategically irrelevant board locations yet still plays moves in those locations.

In this work, we use Go (also known as baduk or weiqi) as a testbed to construct a controlled evaluation set and examine language models' terminology abilities and their relationship to move evaluation. Previous work on Go language models reports that appropriate moves can still be accompanied by inaccurate explanations and misused terminology \citep{ma-etal-2025-mixing}. We therefore examine whether models can accurately relate terminology to board positions and candidate moves as local board configurations and strategic conditions change. 

\begin{figure*}[t]
    \centering
    \begin{subfigure}[t]{0.32\textwidth}
        \centering
        \includegraphics[width=\linewidth]{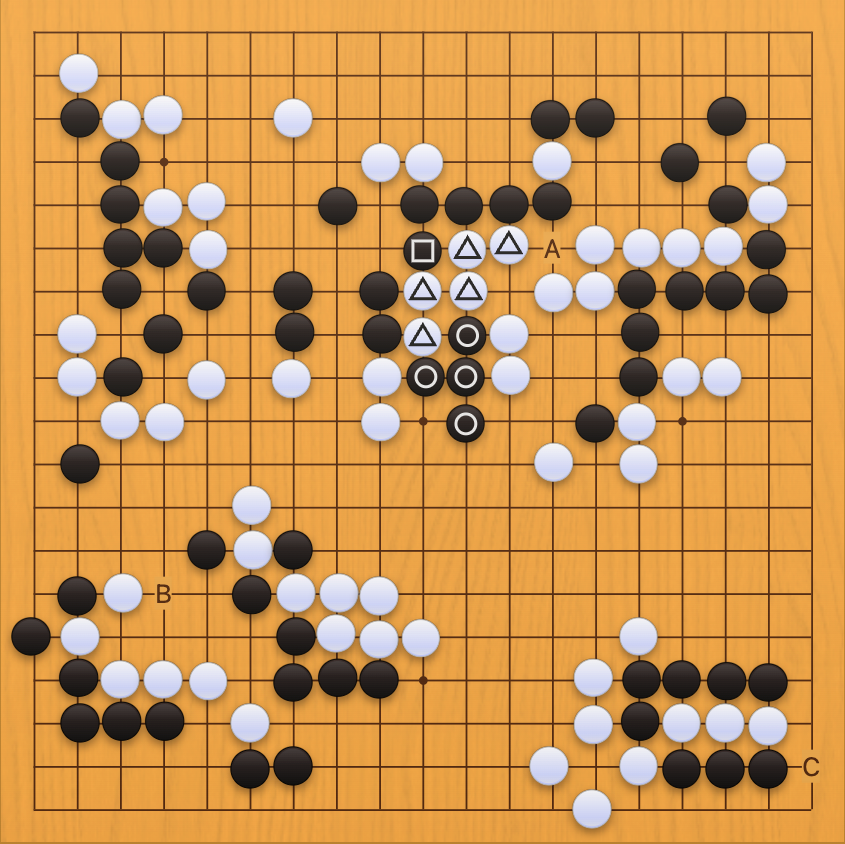}
        \caption{\(X\): \(A\) creates \textit{atari} with a \textit{sente} effect.}
        \label{fig:position-x}
    \end{subfigure}
    \hfill
    \begin{subfigure}[t]{0.32\textwidth}
        \centering
        \includegraphics[width=\linewidth]{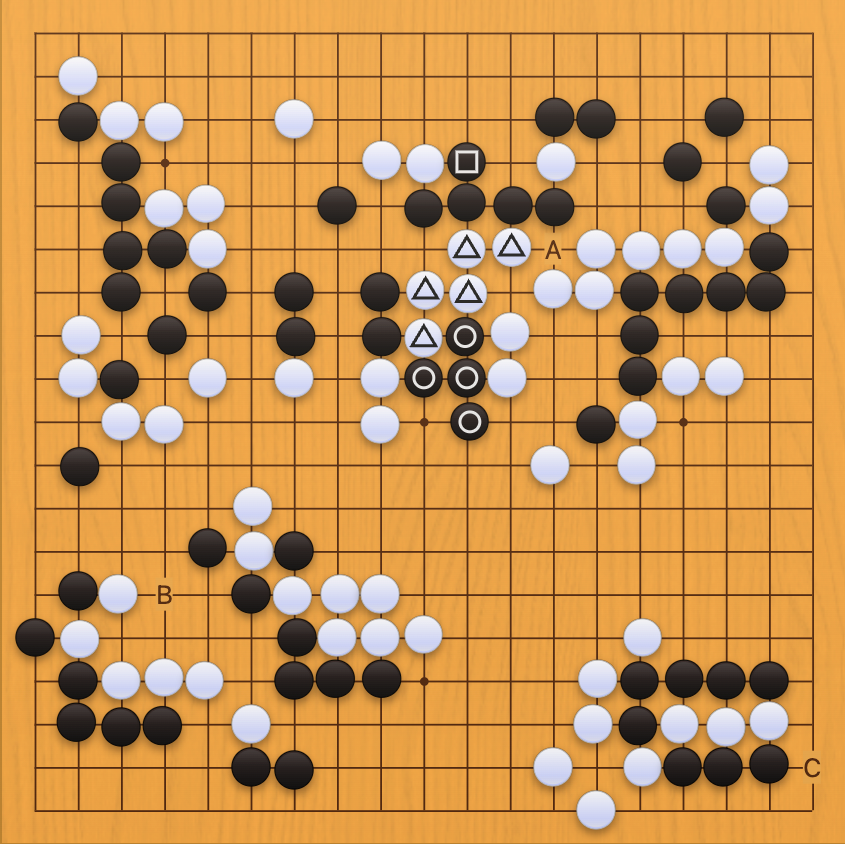}
        \caption{\(Y\): \(A\) no longer creates \textit{atari}.}
        \label{fig:position-y}
    \end{subfigure}
    \hfill
    \begin{subfigure}[t]{0.32\textwidth}
        \centering
        \includegraphics[width=\linewidth]{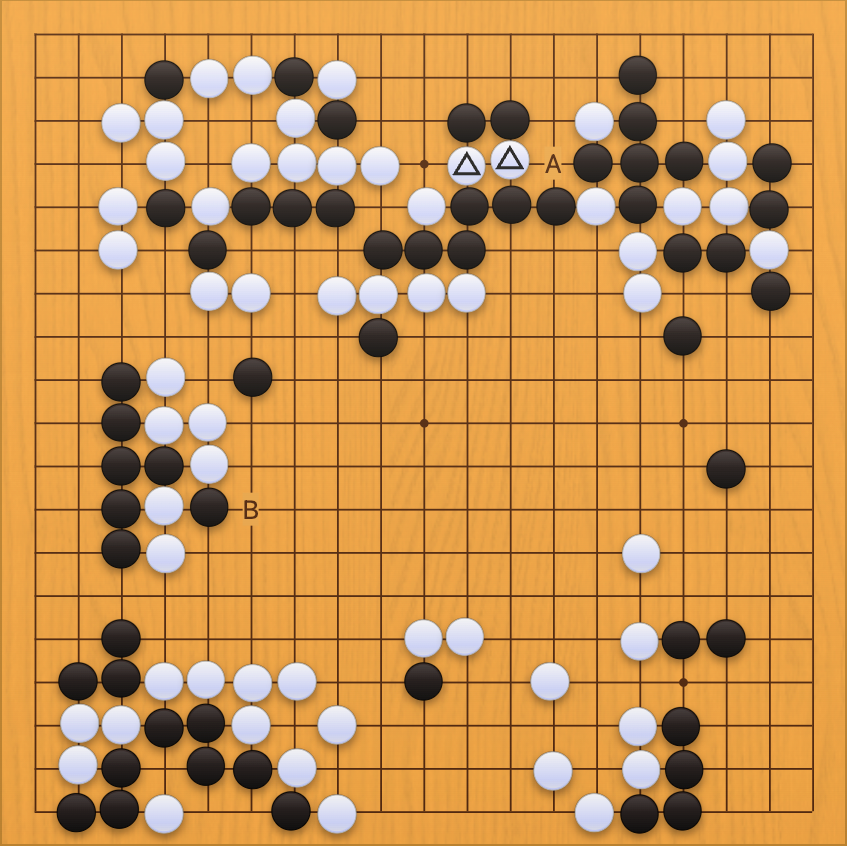}
        \caption{\(Z\): \(A\) creates \textit{atari} without a \textit{sente} effect.}
        \label{fig:position-z}
    \end{subfigure}

    \caption{Examples for the Atari--Sente Contrast Test (\(X/Y\)) and the Non-Sente Atari Test (\(Z\)).}
    \label{fig:atari-counterfactual}
\end{figure*}

Our experiments focus on \textit{atari} and its relationship to \textit{sente}. In Go, an \textit{atari} threatens to capture opposing stones and often forces the opponent to respond, allowing the player to retain the initiative, or \textit{sente}. We construct two complementary tests that vary whether an \textit{atari} applies and whether it has a \textit{sente} effect. We evaluate a fine-tuned version of LoGos, a language model developed to recommend Go moves and generate natural language analyses \citep{ma-etal-2025-mixing}.

Results show that the model does not reliably adjust its terminology judgments in response to local board changes, and that larger groups are associated with lower recognition accuracy. When its win rate predictions indicate a clear preference, it generally distinguishes clearly inferior moves, but struggles to assess the strategic value of \textit{atari} in specific positions. Predicting an \textit{atari} label is also associated with favoring that move, even when this preference contradicts the reference evaluation. We release the evaluation set and construction code to support further research on domain terminology recognition and move evaluation in game-playing language models.\footnote{The code and evaluation set can be found at \url{https://github.com/Iskar-Deng/atari-sente-eval}.}

\section{Experimental Design}
\label{sec:experimental-design}
\subsection{Go Concepts}
\label{sec:go-concepts}

Go is a two-player board game typically played on a \(19 \times 19\) grid. Players alternate placing black and white stones on empty intersections, aiming to control more territory than their opponent. A position consists of the arrangement of stones on the board and the player whose turn it is to move. Stones of the same color connected along horizontal or vertical lines form a group. Empty intersections immediately adjacent to a group along these lines are called liberties. A group is captured and removed from the board when the opponent occupies its last liberty \citep{aga-1991-rules}.

An \textit{atari} in Go is a move that leaves an opposing group with only one liberty, threatening to capture that group on the next turn. The opponent therefore often needs to respond defensively to prevent those stones from being removed from the board. \textit{Sente} generally refers to a move that requires a local response from the opponent, as ignoring it allows the player to follow up and inflict a substantial loss.

Whether an \textit{atari} is \textit{sente} depends on the value of the threatened stones and the alternative moves available on the board. If the cost of ignoring the threat is small, the opponent may choose to sacrifice the threatened stones and play elsewhere.

Figure~\ref{fig:atari-counterfactual} illustrates this relationship. In position \(X\), Black plays \(A\) to create \textit{atari}. If White does not respond, Black can capture the five white stones marked with triangles and rescue the four endangered black stones marked with circles below them. White therefore needs to respond locally, allowing Black to complete a \textit{sente} exchange before playing at the important point \(B\) elsewhere. In this position, both playing \(A\) first and playing \(B\) directly maintain a roughly even position.

In position \(Y\), relocating the black stone marked with a square gives the white group an additional liberty, so \(A\) no longer creates \textit{atari}. White can therefore ignore the move and play elsewhere. Playing \(A\) thus costs Black the opportunity to play \(B\) in time, reducing Black's estimated win rate to approximately \(3\%\), whereas playing \(B\) directly maintains a roughly even position.

Figure~\ref{fig:position-z} illustrates a case where \textit{atari} does not have a \textit{sente} effect. In position \(Z\), Black's move \(A\) threatens two white stones, but White gains more by sacrificing them and playing elsewhere than by saving them. Black's estimated win rate is approximately \(1\%\) after playing \(A\), compared with \(50\%\) after playing at an important point \(B\).

\subsection{Evaluation Tasks}
\label{sec:evaluation-tasks}

Building on the relationship between \textit{atari} and \textit{sente}, we design two tests to examine whether models can accurately identify \textit{atari} and evaluate its value in specific positions. Let \(V(s,m)\) denote the reference value of move \(m\) in position \(s\) from the perspective of the player to move.

\paragraph{Atari--Sente Contrast Test.}
The first test examines whether models adjust their terminology judgments and move evaluations when a local board change removes an \textit{atari} relationship. We construct pairs of positions \(X\) and \(Y\) that differ only in the location of a single stone, as illustrated in Figure~\ref{fig:atari-counterfactual}. \(A\) creates \textit{atari} with a \textit{sente} effect in \(X\) and has a value comparable to \(B\). In \(Y\), \(A\) no longer creates \textit{atari} and is substantially worse than \(B\). We include the same clearly inferior move \(C\) in both positions as a control, giving the following reference relationships:
\[
\begin{aligned}
V(X,A) &\approx V(X,B)>V(X,C),\\
V(Y,B) &>\max\{V(Y,A),V(Y,C)\}.
\end{aligned}
\]

\paragraph{Non-Sente Atari Test.}
The second test examines whether models distinguish the presence of \textit{atari} from its strategic value. We use independent positions \(Z\) where \(A\) creates \textit{atari} without a \textit{sente} effect, as illustrated in Figure~\ref{fig:position-z}. The opponent benefits more from playing elsewhere than from saving the threatened group. For the player to move, an alternative move \(B\) is substantially better:
\[
V(Z,B)>V(Z,A).
\]

We automatically construct the evaluation positions from public Go game records \citep{database-of-go-games}, using rule-based \textit{atari} labels and KataGo's score lead and win rate estimates for selection \citep{wu-2019-accelerating}. The evaluation set contains 1,000 \(X/Y\) pairs and 500 \(Z\) positions. Table~\ref{tab:reference-winrates} reports the mean reference win rates for each candidate move across the evaluation positions. Construction details are provided in Appendix~\ref{app:construction}.

\begin{table}[t]
    \centering
    \small
    \begin{tabular}{@{}lrrr@{}}
        \toprule
        Position & \(A\) & \(B\) & \(C\) \\
        \midrule
        \(X\) & 48.0 & 52.9 & 1.8 \\
        \(Y\) & 4.6 & 48.6 & 1.1 \\
        \(Z\) & 3.8 & 47.9 & --- \\
        \bottomrule
    \end{tabular}
    \caption{Mean KataGo reference win rates (\%) for candidate moves, from the perspective of the player to move.}
    \label{tab:reference-winrates}
\end{table}

\section{Model and Fine-Tuning}
\label{sec:model-training}

We use LoGos-7B as our base model \citep{ma-etal-2025-mixing}. Built on Qwen2.5-7B \citep{qwen-2024-qwen25}, LoGos combines supervised training on Go expert knowledge with reinforcement learning to generate move recommendations and corresponding natural language analyses.

To adapt the model to terminology recognition and win rate prediction for candidate moves, we augment the English Go commentary dataset of \citet{tomlin-etal-2022-understanding} with rule-based terminology labels and KataGo win rate estimates, and jointly fine-tune on terminology, win rates, and human commentary using QLoRA \citep{dettmers-etal-2023-qlora}. When multiple terms apply, we prioritize \textit{atari}. Data processing, annotation methods, input and output formats, and training hyperparameters are provided in Appendix~\ref{app:fine-tuning}.

\section{Results and Analysis}
We evaluate our model on the two tasks described in Section 2.2. For each evaluation position, we extract the move history and board state using the same process as for the training data. Each candidate is evaluated independently using the same input format as in training.

We compare the model's terminology predictions with rule-based labels to assess whether it correctly identifies \textit{atari}. For win rate evaluation, we focus primarily on relative preferences among candidate moves. Previous studies show that small score changes can produce large shifts in Go engines' win rate estimates \citep{egri-nagy-tormanen-2020-derived,rendu-2023-winrate}. In our fine-tuning data, 45.2\% of win rate labels are below 1\% or above 99\%. The model's predictions are also highly concentrated near these extremes. We therefore classify predictions as above or below 50\% to compare position assessments and move preferences rather than numerical win rate accuracy.

\subsection{Terminology Recognition}

The model shows some ability to distinguish \textit{atari} from non-\textit{atari} moves, but struggles with judgments that depend on local changes. As shown in Table~\ref{tab:term-recognition}, it correctly identifies \(B\) as non-\textit{atari} in both \(X\) and \(Y\) for 91.8\% of pairs, and achieves 87.0\% recognition accuracy on \(B\) in \(Z\). For \(A\), however, the most common prediction is \textit{atari} in both \(X\) and \(Y\), accounting for 58.1\% of pairs. Only 14.6\% receive the correct combination of \textit{atari} in \(X\) and non-\textit{atari} in \(Y\). This suggests that the model may recognize local features associated with groups having few liberties, but does not reliably determine whether a move leaves exactly one liberty.

\begin{table}[t]
    \centering
    \small
    \begin{tabular}{@{}llrr@{}}
        \toprule
        Prediction in \(X\) & Prediction in \(Y\) & \(A\) (\%) & \(B\) (\%) \\
        \midrule
        \textit{Atari} & \textit{Atari} & 58.1 & 7.2 \\
        \textit{Atari} & Non-\textit{atari} & \textbf{14.6} & 0.4 \\
        Non-\textit{atari} & \textit{Atari} & 1.5 & 0.6 \\
        Non-\textit{atari} & Non-\textit{atari} & 25.8 & \textbf{91.8} \\
        \midrule
        \multicolumn{2}{@{}l}{Recognition accuracy in \(Z\)} & 59.0 & 87.0 \\
        \bottomrule
    \end{tabular}
    \caption{Terminology recognition results. The upper section reports joint predictions on 1,000 \(X/Y\) pairs; bold marks the correct combinations. The final row reports binary recognition accuracy on 500 \(Z\) positions.}
    \label{tab:term-recognition}
\end{table}

We further examine the relationship between group size and recognition performance by pooling moves \(A\) in \(X\) and \(Z\) that place exactly one group in \textit{atari}. As shown in Table~\ref{tab:group-size}, larger groups are associated with lower recognition accuracy. This may reflect difficulties in identifying larger connected groups and tracking their liberties.

\begin{table}[t]
    \centering
    \small
    \begin{tabular}{@{}lrr@{}}
        \toprule
        Group size & Samples & Accuracy (\%) \\
        \midrule
        2 & 912 & 69.8 \\
        3 & 322 & 64.9 \\
        4 & 109 & 63.3 \\
        \(\geq 5\) & 119 & 58.0 \\
        \bottomrule
    \end{tabular}
    \caption{\textit{Atari} recognition by group size in stones, pooling 962 \(X\) and 500 \(Z\) positions with a single threatened group.}
    \label{tab:group-size}
\end{table}

\subsection{Move Value}

The model shows some ability to distinguish clearly inferior moves. As shown in Table~\ref{tab:move-evaluation}, when its win rate predictions indicate a clear preference, it generally ranks \(C\) below the better alternatives.\footnote{In many comparisons, the model assigns both moves win rates near 0\% or both near 100\%. This may reflect an overly confident assessment of the overall position that obscures differences between moves. We do not count these cases as incorrect rankings.}

\begin{table}[t]
    \centering
    \small
    \begin{tabular}{@{}lrr@{}}
        \toprule
        Reference relation & Coverage (\%) & Cond. acc. (\%) \\
        \midrule
        \(V(X,A)>V(X,C)\) & 24.8 & 97.2 \\
        \(V(X,B)>V(X,C)\) & 21.1 & 90.0 \\
        \(V(Y,B)>V(Y,C)\) & 18.7 & 89.8 \\
        \(V(X,A)>V(Y,A)\) & 11.6 & 88.8 \\
        \(V(Y,B)>V(Y,A)\) & 8.7 & 19.5 \\
        \(V(Z,B)>V(Z,A)\) & 9.0 & 46.7 \\
        \bottomrule
    \end{tabular}
    \caption{Comparisons of candidate move values. Coverage is the proportion of cases where the two predictions fall on opposite sides of 50\%. Conditional accuracy is the proportion of these cases that follow the reference ordering.}
    \label{tab:move-evaluation}
\end{table}

When comparing the value of the same move \(A\) across positions \(X\) and \(Y\), the model achieves a conditional accuracy of 88.8\%. However, this performance does not consistently extend to comparisons between candidate moves within a position. In \(Y\), the model more often incorrectly favors \(A\). In the non-\textit{sente} \textit{atari} positions \(Z\), it correctly favors \(B\) in fewer than half of these comparisons.

Further analysis of terminology predictions shows that labeling \(A\) as \textit{atari} is associated with a greater tendency to favor \(A\) across all three types of position, as shown in Table~\ref{tab:term-value}. This association persists in \(Y\), where \(A\) does not actually create \textit{atari}, and in \(Z\), where the \textit{atari} is not \textit{sente}.

\begin{table}[t]
    \centering
    \small
    \begin{tabular}{@{}llrr@{}}
        \toprule
        Position & Predicted term & Coverage (\%) & Prefer \(A\) (\%) \\
        \midrule
        \(X\) & \textit{Atari} & 10.3 & 96.0 \\
              & Non-\textit{atari} & 8.1 & 40.9 \\
        \midrule
        \(Y\) & \textit{Atari} & 10.9 & 90.8 \\
              & Non-\textit{atari} & 5.4 & 50.0 \\
        \midrule
        \(Z\) & \textit{Atari} & 8.8 & 73.1 \\
              & Non-\textit{atari} & 9.3 & 26.3 \\
        \bottomrule
    \end{tabular}
    \caption{Move preferences grouped by the predicted terminology for \(A\), with coverage calculated within each group.}
    \label{tab:term-value}
\end{table}

\section{Conclusion}
In this paper, we construct two controlled tests centered on \textit{atari} and \textit{sente} in Go to examine how language models recognize domain terminology and relate it to move evaluation. Evaluation of a fine-tuned LoGos model shows that it can identify some \textit{atari} moves, but does not reliably respond to terminology changes caused by local changes in the board. When the model indicates a clear preference, it generally identifies clearly inferior moves, but struggles to determine whether an apparent capture threat makes a move strategically valuable. Furthermore, predicting an \textit{atari} label is associated with favoring that move, even when this preference contradicts the reference evaluation.

Tactical situations in Go can change with small local modifications, requiring models to accurately track the board relationships that determine their judgments. Previous work provides empirical and theoretical evidence of an inductive bias toward low-sensitivity functions in Transformers \citep{bhattamishra-etal-2023-simplicity,hahn-rofin-2024-sensitive}. This bias may pose challenges for game-playing agents that rely entirely on language models to make precise local judgments. Future work on Go AI that generates natural language analyses could explore combining language models with explicit board computations and search to provide accurate local facts and subsequent move sequences. Our controlled tests provide a basis for evaluating whether such approaches improve both local concept recognition and strategic value assessment.

\section*{Limitations}
We follow LoGos in using win rate predictions to evaluate candidate moves. As discussed in Section~4, the sensitivity of win rates and the concentration of model predictions near the extremes limit this evaluation approach. Human players' judgments involve pattern recognition and the integration of local information with the global board context \citep{harre-etal-2012-perceptual}. Evaluating moves through win rate predictions alone therefore makes it difficult to isolate the model's understanding of the relevant local relationships. Future work could incorporate score differences, local board relationships, and subsequent responses as training signals and evaluation criteria to better assess tactical understanding and move evaluation in language-based Go agents.

In addition, we evaluate only a single LoGos model fine-tuned with terminology and commentary supervision. Its performance may reflect the combined effects of the base model, fine-tuning data, and training configuration. We do not compare different training conditions and therefore cannot determine the specific effect of terminology supervision on model performance. Furthermore, we observe an association between terminology predictions and move preferences but do not test whether this relationship is causal. Future work should examine these findings across more models and training conditions and further investigate the sources of the observed errors.

\bibliography{custom}

\appendix

\section{Evaluation Position Construction}
\label{app:construction}

We construct evaluation positions from SGF records collected by \citet{database-of-go-games}. We first identify structural candidates and use score lead differences to verify the required move-value relationships, since win rates can saturate in unbalanced positions. We then apply win rate constraints to select the final evaluation positions. The source website states that the games are in the public domain, and our released evaluation set is intended for research on Go agents.

\subsection{KataGo Configuration}
We use KataGo with Chinese rules and a komi of 7.5 on a \(19 \times 19\) board. Initial screening uses reduced search budgets, while final candidate evaluations use a budget of 1,000 visits per query. Each candidate move is evaluated separately by restricting the root move. Win rates and score leads are reported from the perspective of the player to move; comparisons between defensive responses use the opponent's perspective. All samples retain complete legal move histories, with candidate moves restricted to the first 150 moves.

\subsection{Atari--Sente Contrast Test}
We identify groups with two liberties and candidate moves \(A\) that create \textit{atari}. We construct \(Y\) by changing one historical move's coordinate so that the final boards differ only in the location of one stone and \(A\) no longer creates \textit{atari}. We require the score leads of \(A/B\) in \(X\) to differ by at most 2 points, \(B\) to exceed \(A\) in \(Y\) by at least 4 points, and \(B\)'s score lead to change by at most 3 points across positions. The advantage of \(B\) over \(A\) must increase by at least 3 points from \(X\) to \(Y\). For unbalanced pairs, we append identical moves outside the target region while preserving these relationships. The final win rates of \(A/B\) in \(X\) and \(B\) in \(Y\) must lie within 30--70\%. In \(Y\), \(B\) must exceed \(A\) by at least 20 percentage points, while \(B\)'s win rate may differ by at most 30 percentage points across positions. We select a shared move \(C\) whose win rate is at most 20\% and at least 30 percentage points below \(B\), with a score lead at least 4 points lower, in both positions.

\subsection{Non-Sente Atari Test}
We select positions where \(A\) places exactly one group of at least two stones in \textit{atari}. KataGo's preferred move \(B\) must have a win rate within 30--70\% and a score lead at least 4 points higher than \(A\). After \(A\), the opponent's preferred response must outperform the local saving move by at least 4 points and have a minimum Manhattan distance of 4 from the target stones, \(A\), and the saving point. We verify that the saving move leaves the target group with at least two liberties, whereas after the preferred response elsewhere, the group remains in \textit{atari} and can be captured on the following turn.

\section{Model Fine-Tuning}
\label{app:fine-tuning}

\subsection{Data Annotation}
\label{app:annotation}
Our dataset is created from the Go game records released by \citet{tomlin-etal-2022-understanding}, originally collected from the Go Teaching Ladder, which permits non-commercial digital redistribution of its game records. We process these records to construct training examples consisting of move histories, board states, candidate moves, and human commentary. We extract individual moves associated with comments. Each selected move serves as the candidate move for a separate training example, with its comment used as the output commentary.

The move history is directly extracted from the game record. We convert it to a sequence of numbered moves, each specifying the player color and board coordinate. X and O are for black and white stones respectively. Coordinates follow standard Go notation, with rows 1-19 and columns A-T (excluding I). The board state is built by replaying the moves and represented as a 19 by 19 matrix, where 1 represents black stones, -1 represents white stones, and 0 represents empty positions. Figure~\ref{fig:training-example} shows an example, with the board state abbreviated for readability. The move history and board state in our input follow the textual representation used by LoGos \citep{ma-etal-2025-mixing}.

We annotate moves with Go terminology using a rule-based term matcher, which determines the term by checking the local board state around a move. Each Go term has one or more patterns describing where own stones, opponent stones, and empty intersections should appear relative to the move. These patterns use relative positions so that the same shape can appear anywhere on the Go board. The matcher also checks different rotations of a pattern, since the same concept can appear in different orientations. It can also invert the stone colors, allowing the same pattern to work when either Black or White is playing. We apply this term matcher to annotate the candidate move.

The terms are checked in a predefined order, and the matcher assigns the first term whose pattern matches the board position. Since our evaluation focuses on \textit{atari}, we place it to the front of the matching list, ensuring that \textit{atari} takes precedence when multiple terms can be applied to the same move. To generate win rate labels for the candidate moves, we use KataGo under Chinese rules with a komi of 7.5, consistent with the configuration used for evaluation position construction. The win rate is recorded from the perspective of the player making the candidate move. A higher value indicates a greater probability of winning.

Our training dataset contains 9,398 examples covering the five most common Go terms in the annotated dataset: Stretch, Hane, Tiger's Mouth, Atari, and Connect. Stretch accounts for 22.74\% (2,137 examples), followed by Hane at 21.33\% (2,005), Tiger's Mouth at 20.97\% (1,971), Atari at 17.92\% (1,684), and Connect at 17.04\% (1,601). The distribution is relatively balanced, with each term representing approximately 17–23\% of the training examples.
The win rate labels are concentrated near the extremes, with 45.22\% of examples having win rates below 1\% or above 99\%. Only 17.53\% of examples have win rates between 25\% and 75\%. The mean and median win rates are 44.41\% and 33.00\% respectively.

\begin{figure}[t]
\centering
\begin{minipage}{\columnwidth}
\begin{Verbatim}[fontsize=\footnotesize]
Input:

Given the following Go game state and
candidate move, identify the Go term
that describes the candidate move and
provide commentary explaining the move.

Move history:
1.X-D16 2.O-Q17 3.X-Q4 4.O-F17
5.X-C14 6.O-D4 7.X-L16 8.O-J17
9.X-O16 10.O-Q15 11.X-O14 12.O-R6
13.X-O3 14.O-R8 15.X-D6 16.O-F3
17.X-D9 18.O-D12 19.X-K3 20.O-H3
21.X-D18 22.O-C10 23.X-C9 24.O-B10
25.X-B9 26.O-D14 27.X-C15 28.O-C13
29.X-C5 30.O-C4 31.X-B5 32.O-B4
33.X-F15 34.O-E16 35.X-E15
36.O-D15 37.X-E17 38.O-F16

Board state:
0 0 0 0 0 0 0 0 0 ...
0 0 0 1 0 0 0 0 0 ...
0 0 0 0 1 -1 0 0 -1 ...
0 0 0 1 -1 -1 0 0 0 ...
0 0 1 -1 1 1 0 0 0 ...
...
0 0 0 0 0 0 0 0 0 ...

Candidate move:
X-C16

Output:

Term: Connect
Winrate: 1.73%
Comment: Safe - no need to give your
opponent a place to threaten you
\end{Verbatim}
\end{minipage}

\caption{Example of a training input and
output. The board state is abbreviated.}
\label{fig:training-example}
\end{figure}

\subsection{Fine-Tuning Configuration}
\label{app:fine-tuning-config}
We fine-tune LoGos-7B \citep{ma-etal-2025-mixing}, a 7-billion-parameter model specialized for Go commentary, as our base model. The model is released under the Apache 2.0 license. We fine-tune the model
using QLoRA \citep{dettmers-etal-2023-qlora}
with 4-bit quantization. We apply LoRA to the attention and feed-forward projection layers, with a rank of 16, scaling parameter of 32, and dropout rate of 0.05. We use a learning rate of \(5 \times 10^{-5}\), a batch size of 1 with 8 gradient accumulation steps, and a maximum sequence length of 3,072 tokens. Models are trained for two epochs using supervised fine-tuning on an NVIDIA RTX 4090 GPU with 24 GB memory. Each training run takes approximately 4 hours.

\section{Evaluation Pipeline}
\label{app:evaluation-pipeline}
For each evaluation position, we evaluate candidate moves independently using the same input format as in training. We use greedy decoding with a maximum of 128 new tokens. We extract terminology and win rate predictions from the \textit{Term} and \textit{Winrate} fields in the generated text. For terminology evaluation, we group predicted terms other than \textit{atari}, such as Stretch or Connect, as non-\textit{atari}, without assessing whether these alternative terms are correct. Win rate predictions are interpreted from the perspective of the player making the candidate move. We save the extracted predictions and reference values in JSON format for subsequent analysis.

\end{document}